\documentclass[10pt, conference]{IEEEtran}
\usepackage[left=0.625in,right=0.625in,top=0.75in,bottom=0.99in]{geometry}
\IEEEoverridecommandlockouts
\usepackage{cite}
\usepackage{amsmath,amssymb,amsfonts}
\usepackage{graphicx}
\usepackage{color}
\usepackage{subfig}
\usepackage{stackengine}
\usepackage{comment}
\usepackage[normalem]{ulem}
\usepackage{tabularx}
\usepackage{bm}
\usepackage{dsfont}
\usepackage{mathtools}
\usepackage{lettrine}
\usepackage[font=scriptsize]{caption}

\usepackage[colorlinks, citecolor=blue]{hyperref} 

\usepackage[flushleft]{threeparttable} 
\usepackage{booktabs}

\usepackage{textcomp}
\usepackage{xcolor}

\usepackage[linesnumbered, ruled]{algorithm2e}
\SetKwRepeat{Do}{do}{while}%

\SetCommentSty{mycommfont}
\usepackage{algpseudocode}

\def\BibTeX{{\rm B\kern-.05em{\sc i\kern-.025em b}\kern-.08em
    T\kern-.1667em\lower.7ex\hbox{E}\kern-.125emX}}
    
\usepackage{orcidlink}

\usepackage{xspace}
\usepackage{ifthen}
\usepackage[inline]{enumitem}

\usepackage{algcompatible}

\newboolean{showcomments}
\setboolean{showcomments}{true}
\ifthenelse{\boolean{showcomments}}
{
}

\usepackage{subfig}
\usepackage{fancyhdr}

\begin{document}

\title{\huge Leveraging Multimodal-LLMs Assisted by  Instance Segmentation for Intelligent Traffic Monitoring}

\author{
\IEEEauthorblockN{Murat Arda Onsu$^1$, Poonam Lohan$^1$, Burak Kantarci$^1$, Aisha Syed$^2$, Matthew Andrews$^2$, Sean Kennedy$^2$}\\
\IEEEauthorblockA{\textit{$^1$University of Ottawa, Ottawa, ON, Canada}\\
\textit{$^2$Nokia Bell Labs, 600 March Road,
Kanata, ON K2K 2E6, Canada}\\
$^1$\{monsu022, ppoonam, burak.kantarci\}@uottawa.ca,~$^2$\{aisha.syed, matthew.andrews, sean.kennedy\}@nokia-bell-labs.com}
}

\maketitle
\begin{abstract}
A robust and efficient traffic monitoring system is essential for smart cities and Intelligent Transportation Systems (ITS), using sensors and cameras to track vehicle movements, optimize traffic flow, reduce congestion, enhance road safety, and enable real-time adaptive traffic control. Traffic monitoring models must comprehensively understand dynamic urban conditions and provide an intuitive user interface for effective management. This research leverages the LLaVA visual grounding multimodal large language model (LLM) for traffic monitoring tasks on the real-time Quanser Interactive Lab simulation platform, covering scenarios like intersections, congestion, and collisions. Cameras placed at multiple urban locations collect real-time images from the simulation, which are fed into the LLaVA model with queries for analysis. An instance segmentation model integrated into the cameras highlights key elements such as vehicles and pedestrians, enhancing training and throughput. The system achieves 84.3\% accuracy in recognizing vehicle locations and 76.4\% in determining steering direction, outperforming traditional models.

\end{abstract}



\section{Introduction} \label{sec:1}

Efficient traffic monitoring is crucial to increase road throughput, reduce congestion, and minimize collision risk in smart cities and Intelligent Transportation Systems (ITS). In an urban transportation system, where pedestrians, cars, and public transportation interact at intersections and other conflicting locations, traffic monitoring systems must provide both safety and efficiency in the environment. Through intelligent decision-making, ITS can adjust to changing traffic conditions and deploy strategies to ease congestion, for example, providing alternative routes for vehicles and sending alerting signals to the driver to reduce the risk of collisions. With advanced artificial intelligence (AI) modules, the ITS capabilities have been significantly enhanced in terms of adaptive traffic monitoring under challenging conditions \cite{4}. 

Large Language Model (LLM), a language model with a vast number of pre-trained parameters, trained on massive volumes of data from diverse sources, offers promising potential in decision-making and logical reasoning in complex tasks while providing a human interface for clarity. An ITS and traffic monitoring model can respond to specific situations more quickly and intelligently by utilizing LLMs. Moreover, these models are capable of handling data effectively and producing predictions and reactions that are appropriate for the given situation in the context of traffic monitoring \cite{5}.

The vehicular environment involves textual, image, and video data, necessitating a multimodal model capable of handling diverse datasets. Researchers show that data-centric LLM architectures are effective for traffic monitoring and decision-making \cite{6}. Thus, the Large Language and Vision Assistant (LLaVA) model is adopted, an end-to-end multimodal model integrating a vision encoder with an LLM for visual and language understanding \cite{9}. It enhances LLMs by aligning image features with text. However, using a pre-trained LLaVA model directly may generate excessive or incomplete information from camera captures. To address this, our work fine-tunes the LLaVA model with high-quality data consisting of instance-segmented images and their captions to ensure concise and relevant outputs.

In this research, we fine-tune the pre-trained LLaVA model using Low-Rank Adaptation (LoRA) to specific traffic monitoring tasks with our custom image data and textual descriptions collected from the Quanser Interactive Lab Simulation environment, which is an advanced virtual simulation environment designed for education and research in engineering, robotics, and vehicular tasks \cite{quanser_python_api_2024}. To emphasize the critical features of the image data, such as vehicles and pedestrians, the real-time instance segmentation model \cite{17} is utilized to capture these components and provide insights into the LLaVA model. The main contributions of this work are summarized as follows:

\begin{itemize}
    \item We create a high-quality and diverse dataset using Quanser Interactive Lab to fine-tune a multimodal LLM, incorporating a wide range of scenarios such as traffic congestion, collisions, platooning, pedestrian crossings, etc.
    \item We fine-tune the multimodal LLM model, LLaVA, an advanced framework integrating visual and textual inputs, using LoRA on our custom dataset, and employ it for traffic monitoring tasks. To enhance the model's adaptability to diverse camera deployments, instead of having it memorize each street and road name for each camera perspective, we fine-tune the model on generic aliases associated with the real location names. A separate, externally maintained lookup table then maps these aliases to the corresponding real-world locations.
    \item We utilize a real-time instance segmentation model, YOLOv11, for emphasizing important elements in the collected images to assist the multimodal LLM with traffic analysis and further improve its output.
\end{itemize}

The rest of the paper is organized as follows. Section \ref{sec:2} presents related works. Section \ref{sec:methodology} describes the methodology, including data collection, image segmentation to highlight critical features, and multimodal LLM for traffic monitoring tasks. Section \ref{sec:results} explains the training process and numerical results with visual examples, and Section \ref{sec:conclusion} concludes the paper.

\section{Related Works} \label{sec:2}

The application of LLMs to multimodal data is a highly active and rapidly developing area of research. In study \cite{2}, researchers examine how well these models, including VideoLLaMA-2, GPT-4o, and others, can analyze real-world and synthetic traffic videos to answer complex queries related to traffic conditions, vehicle movements, and incidents. The evaluation is conducted using a framework that measures model accuracy, relevance, and consistency across different query categories, such as basic detection, temporal reasoning, and compositional queries. The research \cite{7} explores the integration of multimodal LLM models with street view images (SVIs) to assess urban safety perception. This study leverages state-of-the-art multimodal LLMs, such as GPT-4 Vision (GPT-4V), to automatically rank safety scores for different urban environments based on SVIs. The researchers constructed a benchmark dataset, via Baidu Maps, using human-annotated safety scores and demonstrated that multimodal-LLM-generated scores closely align with human perception. Moreover, the work in \cite{8} investigates the use of multimodal-LLM, such as Gemini-Pro Vision 1.5, to detect traffic safety critical events such as sudden stops, lane changes, or crossing pedestrians that could lead to dangerous situations if not managed correctly in autonomous driving environments. 

Another research \cite{11} introduces TrafficVLM, a controllable visual-language model (VLM) designed for dense video captioning in traffic monitoring scenarios, which generates detailed, multi-phase descriptions of both vehicles and pedestrians in various traffic conditions. It processes overhead camera views, capturing fine-grained spatial and temporal details, which makes it a valuable tool for urban surveillance, ITS, and autonomous driving applications. Also, iLLM-traffic signal control (TSC) is introduced in \cite{12} to address the limitation of real-world challenges such as packet loss, communication delays, and rare events by employing a dual-step decision-making process where a reinforcement learning (RL) agent first makes an initial decision, which is then refined by an LLM that incorporates broader environmental context, logical reasoning, and missing information. Furthermore, researchers in \cite{13} explore the integration of LLMs in Adaptive Traffic Signal Control (ATSC) by applying two LLM-based traffic controllers: Zero-Shot Chain of Thought (ZS-CoT) and a Generally Capable Agent (GCA)-based controller. The ZS-CoT controller relies solely on the capabilities of the actor agent to operate without prior specific training on tasks, which resembles an open-loop feedback control system. It uses LLMs for reasoning and decision-making without prior fine-tuning. The GCA-based controller on the other hand improves upon this by integrating feedback from real-time traffic conditions to refine its decision-making process.

TrafficGPT is introduced in \cite{14} as a novel framework that integrates LLMs with Traffic Foundation Models (TFMs) to enhance traffic management and decision-making and address the numerical data processing and real-time interaction challenges of the LLMs. This framework bridges this gap by combining multimodal traffic data sources with domain-specific models that handle traffic analysis, visualization, and control. Moreover, researchers in \cite{15} propose a framework, SeeUnsafe, that integrates multimodal LLMs to enhance traffic accident analysis from video data. It shifts manual post-processing of raw vision-based paradigm by enabling a conversational, AI-driven approach that automates video classification and visual grounding to classify the condition as normal, near-miss, and collision, extracting structured safety insights. Additionally, in study \cite{16}, ChatSUMO, an LLM-based assistant designed to automate traffic scenario generation within the Simulation of Urban MObility (SUMO) platform, is utilized, which leverages Meta's Llama 3.1 model to create, modify, and analyze traffic simulations using natural language commands, eliminating the need for coding expertise. ChatSUMO supports dynamic adjustments such as modifying road networks, optimizing traffic lights, and altering vehicle types, facilitating more flexible and interactive traffic planning via text commands. 

In contrast to the majority of related works, which mainly concentrate on classifying specific, isolated traffic scenarios (collisions, intersection behavior, single-vehicle driving), this research, this research aims to provide comprehensive traffic monitoring. Using a fine-tuned, visually grounded multimodal LLM, and help with instance segmentation, we analyze the entire traffic environment captured by multiple cameras, considering a wider range of use cases with multiple vehicles, including traffic congestion, collisions, platooning, and pedestrian crossings, incorporating diverse elements such as intersections, roundabouts, long roads, pedestrians, crosswalks, and traffic signs.

\begin{figure*} [t]
    \centering
    \includegraphics[width=0.90\linewidth]{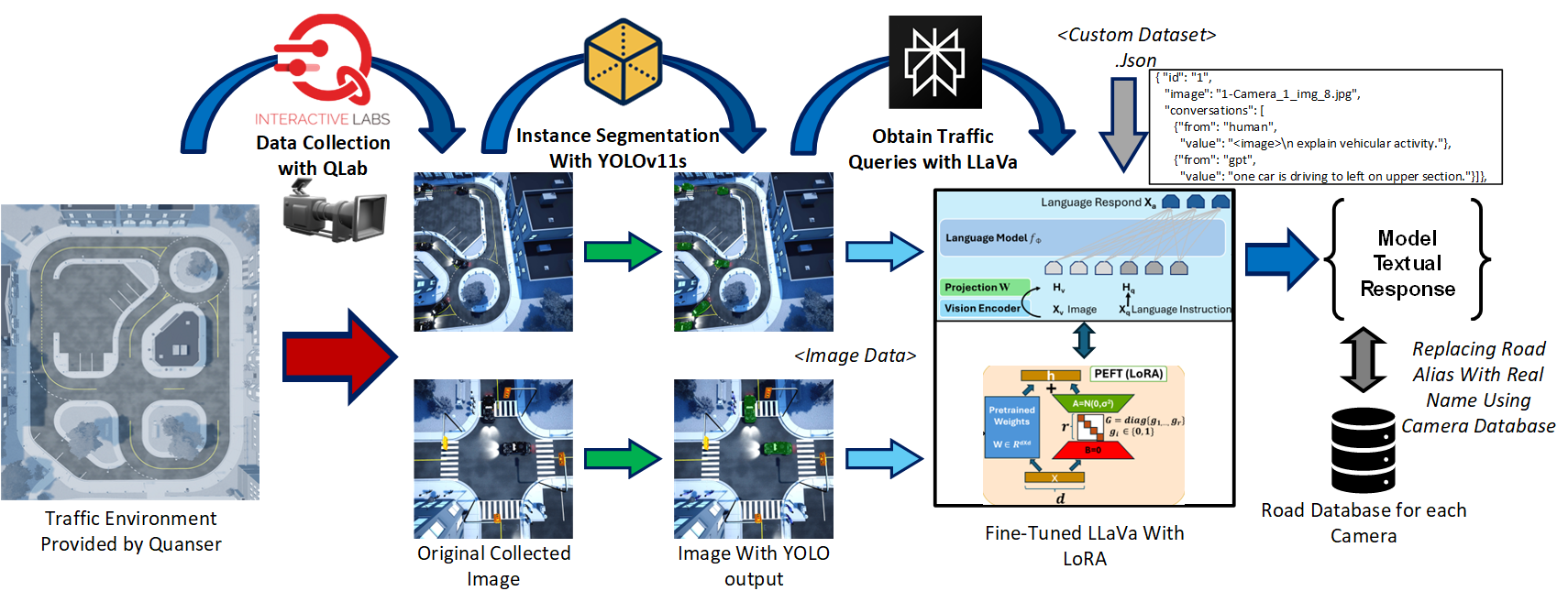}
    \caption{Overall Workflow of our Traffic Monitoring Pipeline Including Data Collection, Instance Segmentation and LLaVA Model.}
    \label{fig: workflow}
\end{figure*}

\section{Methodology} \label{sec:methodology}

In this research, the main aim is to leverage a multimodal LLM that can utilize both image and textual data to give proper responses to queries related to traffic monitoring. Therefore, it is essential to set up an environment where various driving scenarios can be established and data, both textual and image, is collected in a high-quality and diverse way. To improve the performance of the LLM model, essential objects in the data should be emphasized, especially on images, since language models might not be as efficient on this kind of data as much as textual data. To address this problem, the image segmentation method is applied to each image collected from the environment, which highlights the objects, before sending them to the LLM. After the data is ready, the multimodal LLM model, LLaVA, is fine-tuned on the custom dataset to perform traffic monitoring tasks. To improve the flexibility of LLaVA model deployment on different cameras, instead of memorizing roads' names, we ensure that it tries to understand roads' locations on the image using generic aliases and then use an externally maintained database to extract the real names of the generic aliases. The overall workflow of the our methodology, including data collection, instance segmentation, and multimodal LLM, is shown in \figurename \ref{fig: workflow}

\subsection{Data Collection} \label{sec:3}

\begin{figure} [t]
    \centering
    \includegraphics[width=0.90\linewidth]{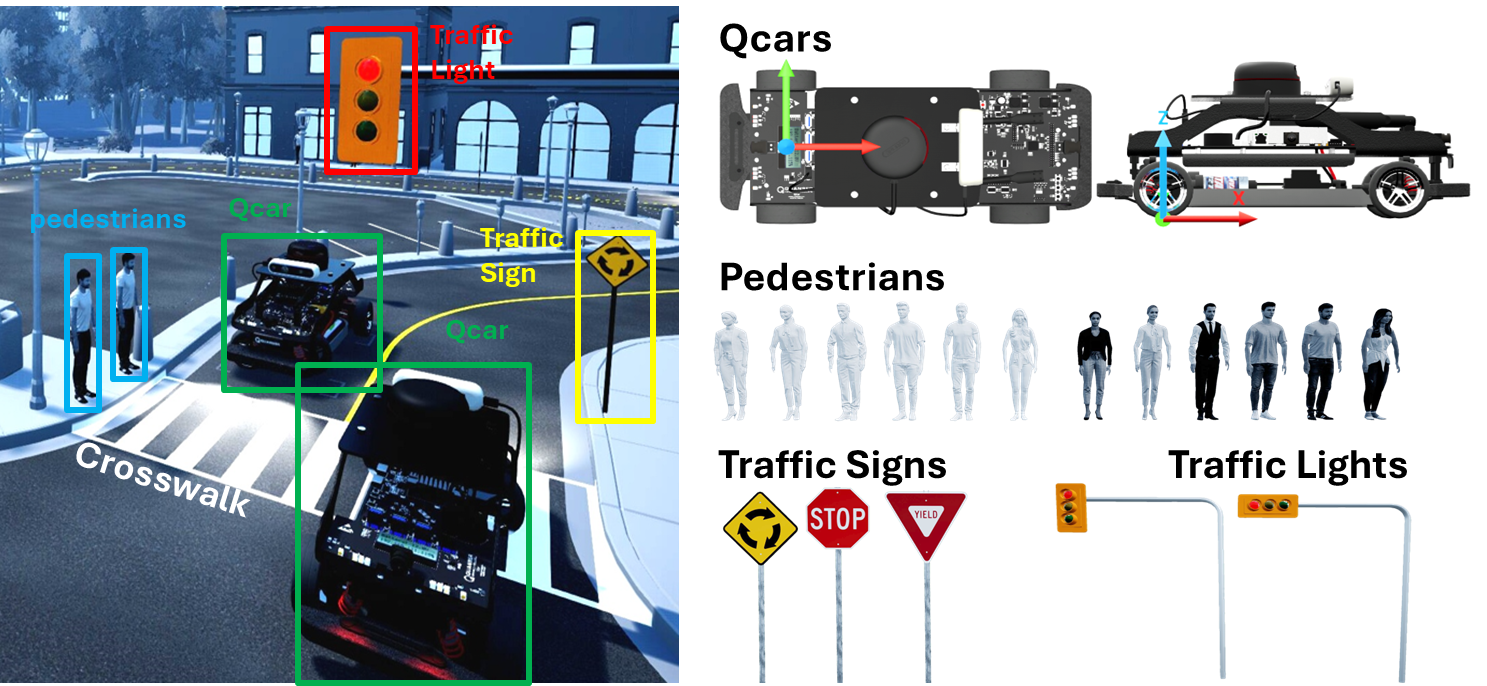}
    \caption{Quanser Interactive Lab Simulation Runtime Captured Image (On Left) and Components (On Right) for Data Collection}
    \label{fig:img1}
\end{figure}

Traffic monitoring tasks using multimodal LLMs require extensive and high-quality image data in their textual format. Since it is challenging to perform experiments with diverse scenarios in the real world, such as creating accidents and congestion, with different camera perspectives over and over again, so we choose a simulation platform, the Quanser Interactive Labs (QLabs), for the initial model training \cite{quanser_python_api_2024}. QLabs allows users to experiment, design, and test algorithms in a safe and controlled virtual environment, making it ideal for areas like control and monitoring systems by providing realistic and flexible traffic components such as roads, vehicles, pedestrians, traffic signs, lights, crosswalks and so on which can be seen in \figurename \ref{fig:img1}.

\begin{figure} [t]
    \centering
    \includegraphics[width=0.90\linewidth]{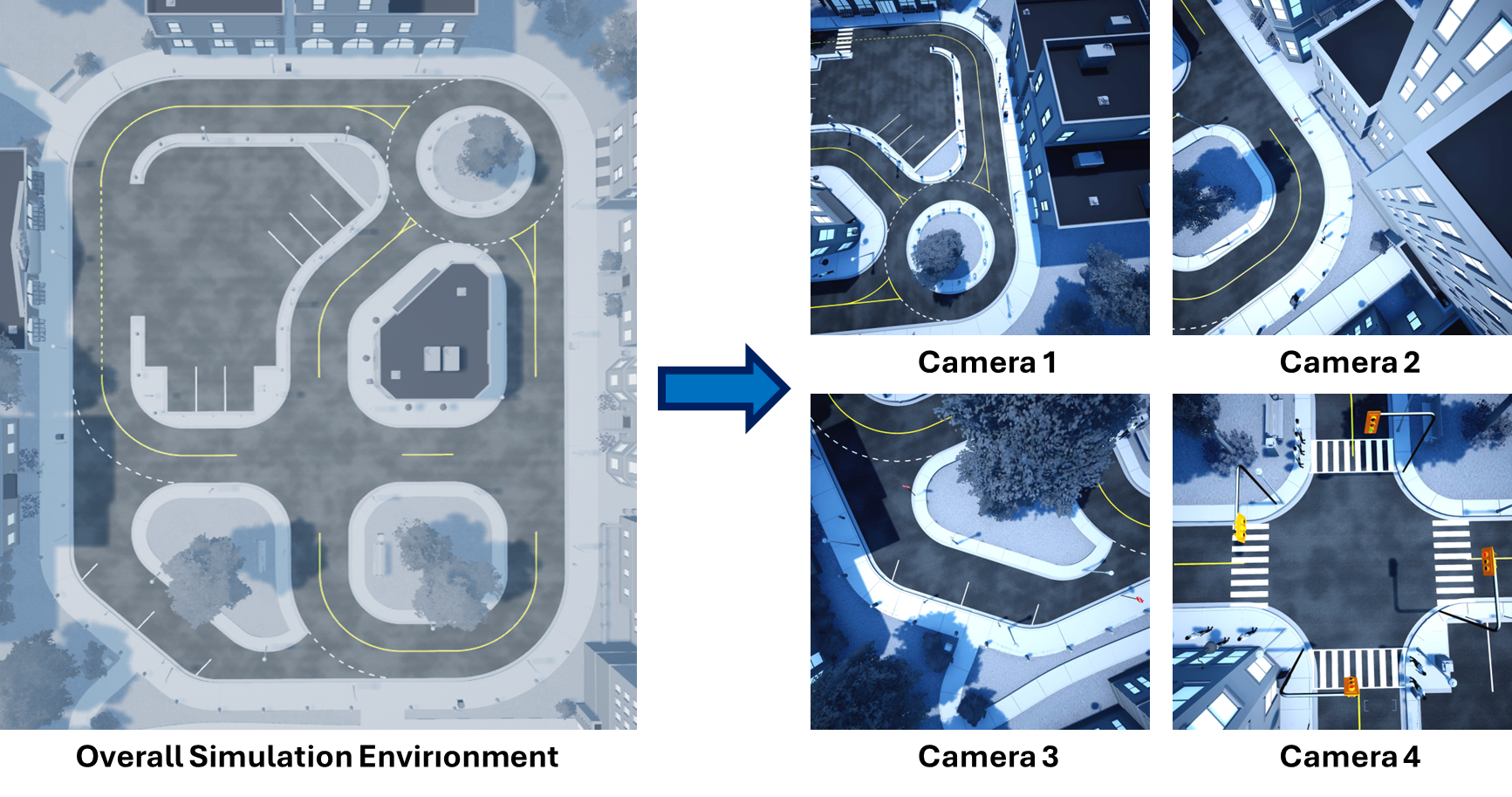}
    \caption{Overall Simulation Environment and Attached Camera view during the Runtime}
    \label{fig:img2}
\end{figure}

According to \figurename \ref{fig:img1}, the left half shows the captured images from simulation runtime, and the right half explains the components of the Qlab, which can be generated as much as required, placed, and moved anywhere in the simulation environment. There are three different traffic signs, Stop, Yield, and Roundabout, and 12 different people figures (6 for low and 6 for high detailed people model). One of the most important components of the simulation car, namely Qcar, includes various sensors such as LiDAR and radar, and five different cameras, front, back, left, right, and top camera, which can also capture depth images. The researcher can adjust the location and rotation of the traffic lights, signs, and crosswalks, and pedestrians and vehicle movements, speed, and path can be defined before the simulation run. This simulation platform provides AI model integration for several studies such as traffic sign detection, intersection study, the collision of car to car or car to pedestrians, and so on. Its compatibility with widespread programming environments such as MATLAB/Simulink and Python enables seamless integration with existing workflows, fostering hands-on learning and innovative research in engineering and AI fields. The simulation environment can be seen in \figurename \ref{fig:img2}.

The left part of \figurename \ref{fig:img2} shows the overall empty environment with no weather condition set. Qlab simulation platform includes camera components that can be attached anywhere and run during the simulation runtime for image collection with configurable resolution and rate. In our setup, four different cameras are attached to various places in the environment that are not overlapped, and the combination represents the overall environment, as can be seen on the right side of the \figurename \ref{fig:img2}. In this simulation, data is collected every 5 ms with 1024x1024 resolution. 30 different driving scenarios are established with various numbers of Qcars, including intersections, roundabouts, 4-way junctions, one-way roads, and so on. A total of 800 image data with textual descriptions are collected for model training. Moreover, 7 experiments among overall scenarios include collision events to teach the model to check the collision condition or likelihood of collision probability during training.    

\subsection{Instance Segmentation} \label{sec:4}

Although LLMs can perform outstandingly on textual data but even multimodal LLMs may struggle in the visual context, especially in very specific scenarios such as traffic monitoring tasks in an unfamiliar environment. This is because multimodal models such as LLaVA are trained primarily on general tasks with general daily life images and Qlab images and output that are required is highly specific to traffic environments so the model can struggle. Thus, during model training, it is essential to emphasize the critical features on the images, such as traffic lights, pedestrians, crosswalks, and vehicles, so that the multimodal LLM model can take the clues to which place it should consider before answering the prompt.

In an urban traffic scenario, the environment consists of static and dynamic components in terms of location, where static components' location does not change during the simulation runtime, while dynamic components typically move to different places with various velocities. Traffic signs, roads, and traffic lights are considered static components, and their pixel-based location can be set on the image before the simulation run. Then, these pixels can be highlighted directly after the camera captures them. However, dynamic components such as cars and pedestrians can actively move in the environment, so it is not appropriate to define their image location before the run. Therefore, it is essential to use proper object detection and instance segmentation models to capture these components in real-time with high accuracy.

To solve this problem, in our setup, the YOLO (You Only Look Once) model, which is widely used in computer vision for real-time object detection and, in some versions, instance segmentation, is integrated into the camera to get the exact location of the dynamic component. YOLO provides rapid speeds and detects objects across the entire field with a single pass. It is based on convolutional neural networks (CNNs), and during its workflow, it divides images into several grids for object detection and predicts bounding boxes, class probabilities, and confidence scores for objects within each grid cell \cite{17}.

To be specific, in this research, YOLOv11, an advanced YOLO-based architecture, is adopted for instance segmentation, leveraging its enhanced speed, accuracy, and ability to delineate individual objects. The model incorporates key architectural components like the Spatial Pyramid Pooling - Fast (SPPF) module for multi-scale feature extraction and the Cross Stage Partial with Spatial Attention (C2PSA) module to enhance spatial focus. These improvements enable YOLOv11 to capture fine details efficiently while maintaining computational efficiency \cite{10}\cite{1}. We select the YOLOv11 small (YOLOv11s) variant for segmenting two dynamic object categories: vehicles and pedestrians. Highlighting these critical aspects on the images helps improve the output of the LLaVA model.


\subsection{Traffic Monitoring using Multimodal-LLM} \label{sec:5}

We fine-tune the LLaVA model with LoRA (Low-Rank Adaptation) for traffic monitoring queries. LLaVA has a multimodal architecture designed to integrate vision and language for tasks requiring a comprehensive understanding of both images and text. It extends the capabilities of LLMs by incorporating a visual encoder, enabling it to process and align visual features with linguistic reasoning.

The architecture of the LLaVA model combines a vision encoder, a projection layer, and a language model to generate language responses based on both image and textual inputs. The process begins with an image input that is processed by a vision encoder to extract image features. These visual embeddings are then passed through a projection layer, aligning them with the language model’s embedding space to produce the visual representation. Simultaneously, a textual query or instruction is tokenized and embedded, generating textual representations. The language model, typically a pre-trained large-scale transformer like LLaMA, receives both visual and textual embeddings as input. Using cross-attention mechanisms, it generates a coherent language response that aligns with the visual context. This architecture enables LLaVA to perform tasks such as image-based question answering, caption generation, and multimodal reasoning, making it highly effective for applications in AI-assisted vision-language understanding.

This research utilizes LLaVA 1.5 to process traffic queries in an urban environment, enhancing multimodal vision-language tasks through visual instruction tuning \cite{3}. LLaVA 1.5 improves visual token extraction and fine-grained feature representation, refining spatial understanding. The projection layer has been optimized for better multimodal alignment, while the transformer-based language model benefits from fine-tuning on diverse instruction datasets, enhancing its ability to interpret visual inputs, answer multimodal queries, and generate contextually accurate responses.

\begin{figure} [t]
    \centering
    \includegraphics[width=0.99\linewidth]{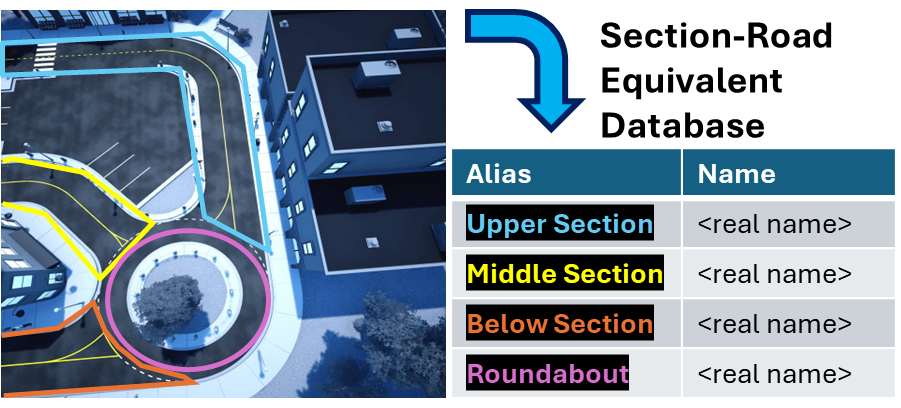}
    \caption{Camera View With Highlighted Road Sections (Left) and Roads' Name Database (Right)}
    \label{fig:img5}
\end{figure}

For fine-tuning LLaVA 1.5, the dataset collection includes image data generated using the Qlab simulation runtime by four different cameras attached to various places in the environment (as mentioned in section \ref{sec:3}). These images are then sent to the YOLOv11s model for emphasizing the important aspects to assist LLaVA model (as mentioned in section \ref{sec:4}). In addition to that, textual data captions for each image are automatically collected in Qlab by taking the coordinates of the vehicles and pedestrians, provided by the simulation tool, to locate their places, such as which road or street they used during the simulation. However, it must be noted that road or street names in each place can vary, and forcing the AI model to memorize the names of these locations makes the training challenging and provides inflexibility in the case of using the same model in a different camera, which requires the model to be trained with new street/road name data. Therefore, these locations on the images and directions of components should be trained to the model in general terms, which can be suitable for each camera caption. To make the model's response realistic for the real world, each general content of the model response output in terms of location should have an equivalent, and these associations between location aliases and real names should be stored in each camera's database. The workflow of this procedure can be seen in \figurename \ref{fig:img5}

As can be seen in \figurename \ref{fig:img5}, instead of writing each road a specific name, their locations are divided into sections. After the model generates an output, these section names in the model response are replaced with actual values, road names, in the camera's database. This methodology, after being appropriately trained, allows the multimodal LLM model to be integrated into any camera and provides flexibility for environmental changes.

\section{Training and Numerical Results} \label{sec:results}

\begin{figure} [t]
    \centering
    \includegraphics[width=0.90\linewidth, height=5cm]{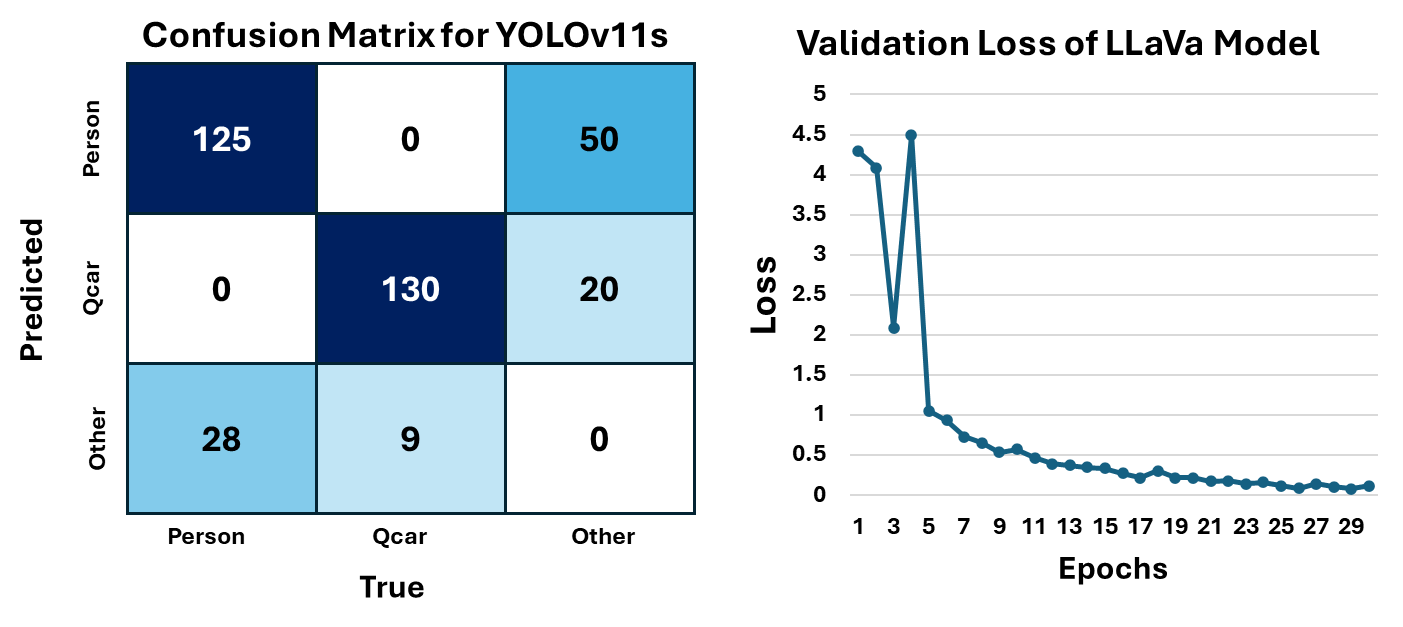}
    \caption{Training and Validation Results for YOLOv11s and LLaVA Model}
    \label{fig:img6}
\end{figure}

The training procedure begins with training the YOLOv11s object detection and image segmentation model using 200 images captured by cameras from several simulation scenarios with batch size of 4 and with 16 iterations. After the image segmentation model is trained sufficiently, it is integrated into each camera, and new experiments are performed for LLaVA model training, including 23 standard driving and 7 collision scenarios providing 800 new camera captions. LoRA, which is an efficient fine-tuning approach for LLaVA models is used by us to adapt LLaVA 1.5 to our specific traffic datasets and we train it for 30 epochs. Training and validation results for YOLOv11s and LLaVA model can be seen in \figurename \ref{fig:img6}.

According to results shown in \figurename \ref{fig:img6}, the YOLOv11s model performs well in identifying Qcars, where 130 samples are correctly classified, while there are 9 instances of Qcars are misclassified as other objects and 20 other objects are misclassified as Qcars. Qcars and pedestrians are not confused by the instance segmentation model, but 50 other objects are misclassified as persons, and 28 persons are misclassified as other objects. The Qcar class performed best, with 86.7\% precision, 93.5\% recall, and an F1-score of 89.9\%, showing substantial recognition accuracy. Moreover, for the person class, the model achieved a precision of 71.4\% and a recall of 81.7\%, resulting in an F1-score of 76.2\%, indicating moderate performance. Here, other objects include trees, poles, and benches near roads. Moreover, environmental factors, such as light shining on the car's roof or dark areas, make it difficult for the model to properly detect objects. In our next work, we plan to enhance the model by incorporating additional training for other objects to improve overall detection accuracy.

The validation loss for LLaVA 1.5 training starts high at 4.29 but rapidly decreases in the initial epochs. Although the loss value reaches 4.5 in the 4th epoch, it decreased to 1.05 by the 5th epoch. The decline then becomes more gradual, stabilizing around 0.3 to 0.2 in the middle epochs, indicating consistent learning. In the final epochs, the loss drops below 0.1, with values like 0.0812 and 0.1071, suggesting strong convergence. The steady decrease without major spikes indicates effective training with no severe overfitting, and the final low values confirm that the model has likely reached an optimized state.

\begin{figure} [t]
    \centering
    \includegraphics[width=84mm]{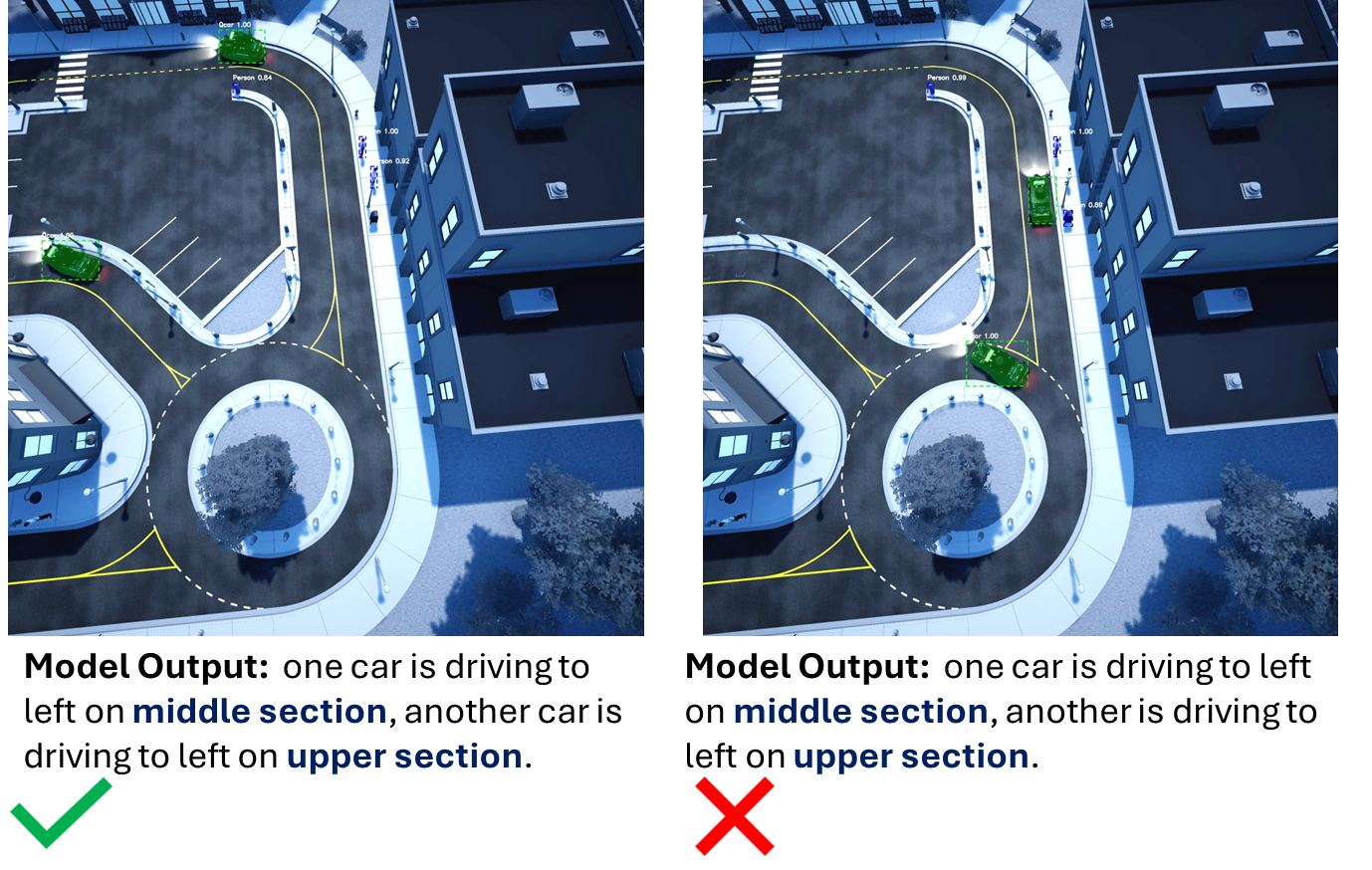}
    \caption{Examples of Correct and Incorrect Outputs of LLaVA Model for Driving Scenario}
    \label{fig:img7}
\end{figure}

The validation scores of LLaVA are evaluated based on car events such as their locations and steering directions in captured scenes. In terms of location-based performance, LLaVA achieves 84.3\% accuracy in correctly identifying car locations, aided by the simplification of location labels by section using the location database, and the use of highlighted objects extracted with YOLO. However, since LLaVA was trained on single-image data, it struggles with accurately determining cars' steering directions. Despite its ability to recognize locations effectively, direction-based validation shows that LLaVA correctly identifies 76.4\% of the cars' steering directions. The results indicate that rather than using the pre-trained LLaVA model directly, fine-tuning with LoRA enhances its performance, leading to a 10.3\% improvement in location identification and a 5.4\% improvement in steering direction recognition. An example output of LLaVA is shown in \figurename \ref{fig:img7}.   

\begin{figure} [t]
    \centering
    \includegraphics[width=80mm]{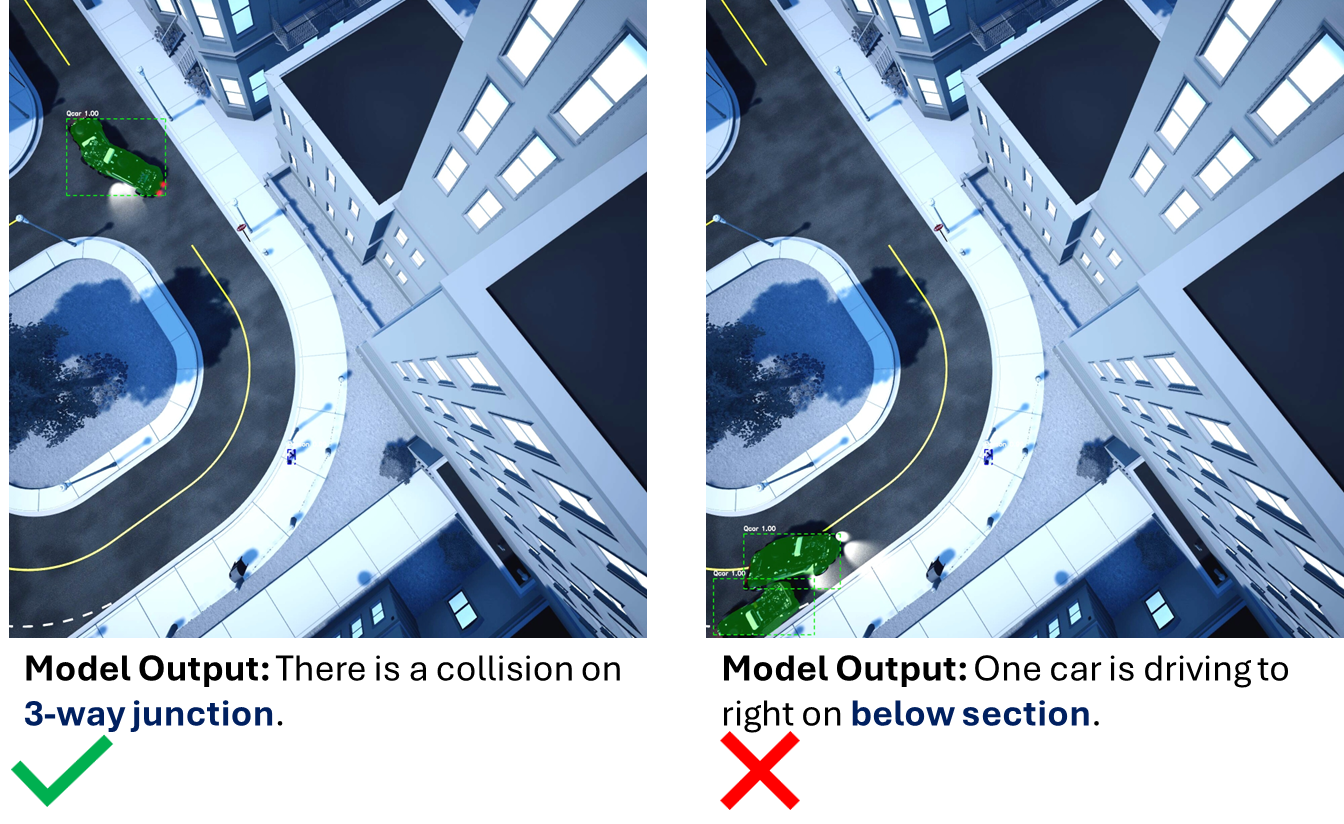}
    \caption{Examples of Correct and Incorrect Outputs of LLaVA Model for Collision Scenario}
    \label{fig:img8}
\end{figure}

The left side of \figurename \ref{fig:img7} presents an example of the correct explanation generated by the LLaVA model for the given image and query, "explain the vehicular activity." In contrast, the right side demonstrates an incorrect response. Text highlighted in blue within the responses will be replaced with the corresponding real names from the camera database, as detailed in section \ref{sec:5}. In the incorrect response, the first car's location is misidentified, it should be on the roundabout rather than in the middle section. Additionally, the second car's steering direction is incorrect, it should be moving upward. Collision scenario examples are illustrated in \figurename \ref{fig:img8}. While the LLaVA model correctly identifies a collision at the three-way junction in the left image, it fails to detect a collision event in the right image, despite accurately describing the location and steering direction of the vehicles.  

\section{Conclusions and Future Work} \label{sec:conclusion}

In this research, urban traffic monitoring using the multimodal LLM called LLaVA has been studied within an advanced simulation platform, Qlab. The primary objective has been to observe the environment and generate accurate responses from the LLaVA model based on given queries and images to extract relevant traffic information. While Qlab has provided an efficient environment and tools for collecting both textual and image data for the LLaVA model, a real-time instance segmentation model has been employed to further enhance LLaVA image analysis. Essential objects are highlighted on the images before they are processed by the LLaVA model with textual queries. The results have demonstrated that the LLaVA model has achieved 84.3\% accuracy in identifying vehicle locations and 76.4\% accuracy in determining steering directions.

These findings have indicated that this research can be extended to more complex simulation environments and challenging queries. Moving forward, this model and methodology will serve as the foundation for further developments. Future work will focus on simulating larger-scale traffic scenarios with an increased number of vehicles, incorporating more advanced challenges such as platooning, collision avoidance, and alternative route planning. Additionally, efforts will be made to enhance the model’s resilience to environmental variations while introducing new methodologies to improve training efficiency, reduce computational resource consumption, and accelerate convergence time.


\section*{Acknowledgment }\label{Section6}
This work is supported in part by the MITACS Accelerate Program, and in part by the Natural Sciences and Engineering Research Council of Canada (NSERC) CREATE TRAVERSAL program. The authors would like to thank Quanser for their support in the generation of the traffic data via Quanser Interactive Lab (https://www.quanser.com/products/quanser-interactive-labs/).



\bibliographystyle{IEEEtran}

\begin{thebibliography}{10}
\providecommand{\url}[1]{#1}
\csname url@samestyle\endcsname
\providecommand{\newblock}{\relax}
\providecommand{\bibinfo}[2]{#2}
\providecommand{\BIBentrySTDinterwordspacing}{\spaceskip=0pt\relax}
\providecommand{\BIBentryALTinterwordstretchfactor}{4}
\providecommand{\BIBentryALTinterwordspacing}{\spaceskip=\fontdimen2\font plus
\BIBentryALTinterwordstretchfactor\fontdimen3\font minus \fontdimen4\font\relax}
\providecommand{\BIBforeignlanguage}[2]{{%
\expandafter\ifx\csname l@#1\endcsname\relax
\typeout{** WARNING: IEEEtran.bst: No hyphenation pattern has been}%
\typeout{** loaded for the language `#1'. Using the pattern for}%
\typeout{** the default language instead.}%
\else
\language=\csname l@#1\endcsname
\fi
#2}}
\providecommand{\BIBdecl}{\relax}
\BIBdecl

\bibitem{4}
D.~Mahmud, H.~Hajmohamed, S.~Almentheri, S.~Alqaydi, L.~Aldhaheri, R.~A. Khalil, and N.~Saeed, ``{Integrating LLMs with ITS: Recent Advances, Potentials, Challenges, and Future Directions},'' \emph{arXiv preprint:2501.04437}, 2025.

\bibitem{5}
S.~Masri, H.~I. Ashqar, and M.~Elhenawy, ``{Large Language Models (LLMs) as Traffic Control Systems at Urban Intersections: A New Paradigm},'' \emph{arXiv preprint:2411.10869}, 2024.

\bibitem{6}
D.~Le, A.~Yunusoglu, K.~Tiwari, M.~Isik, and I.~Dikmen, ``{Multimodal LLM for Intelligent Transportation Systems},'' \emph{arXiv preprint:2412.11683}, 2024.

\bibitem{9}
H.~Liu, C.~Li, Q.~Wu, and Y.~J. Lee, ``{Visual instruction tuning},'' \emph{Advances in neural information processing systems}, vol.~36, 2024.

\bibitem{quanser_python_api_2024}
\BIBentryALTinterwordspacing
Quanser, \emph{{Quanser Python API}}, 2024. [Online]. Available: \url{https://docs.quanser.com/quarc/documentation/python/getting_started.html}
\BIBentrySTDinterwordspacing

\bibitem{17}
C.~H. Kang and S.~Y. Kim, ``{Real-time object detection and segmentation technology: an analysis of the YOLO algorithm},'' \emph{JMST Advances}, vol.~5, no.~2, pp. 69--76, 2023.

\bibitem{2}
J.~R. Vishal, D.~Basina, A.~Choudhary, and B.~Chakravarthi, ``{Eyes on the Road: State-of-the-Art Video Question Answering Models Assessment for Traffic Monitoring Tasks},'' \emph{arXiv preprint:2412.01132}, 2024.

\bibitem{7}
J.~Zhang, Y.~Li, T.~Fukuda, and B.~Wang, ``{Revolutionizing Urban Safety Perception Assessments: Integrating Multimodal Large Language Models with Street View Images},'' \emph{arXiv preprint arXiv:2407.19719}, 2024.

\bibitem{8}
M.~A. Tami, H.~I. Ashqar, and M.~Elhenawy, ``{Using Multimodal Large Language Models for Automated Detection of Traffic Safety Critical Events},'' \emph{arXiv preprint:2406.13894}, 2024.

\bibitem{11}
Q.~M. Dinh, M.~K. Ho, A.~Q. Dang, and H.~P. Tran, ``{Trafficvlm: A controllable visual language model for traffic video captioning},'' in \emph{Proceedings of the IEEE/CVF Conference on Computer Vision and Pattern Recognition}, 2024, pp. 7134--7143.

\bibitem{12}
A.~Pang, M.~Wang, M.-O. Pun, C.~S. Chen, and X.~Xiong, ``{iLLM-TSC: Integration reinforcement learning and large language model for traffic signal control policy improvement},'' \emph{arXiv preprint:2407.06025}, 2024.

\bibitem{13}
M.~Movahedi and J.~Choi, ``{The Crossroads of LLM and Traffic Control: A Study on Large Language Models in Adaptive Traffic Signal Control},'' \emph{IEEE Transactions on Intelligent Transportation Systems}, 2024.

\bibitem{14}
S.~Zhang, D.~Fu, W.~Liang, Z.~Zhang, B.~Yu, P.~Cai, and B.~Yao, ``{Trafficgpt: Viewing, processing and interacting with traffic foundation models},'' \emph{Transport Policy}, vol. 150, pp. 95--105, 2024.

\bibitem{15}
R.~Zhang, B.~Wang, J.~Zhang, Z.~Bian, C.~Feng, and K.~Ozbay, ``{When language and vision meet road safety: leveraging multimodal large language models for video-based traffic accident analysis},'' \emph{arXiv preprint arXiv:2501.10604}, 2025.

\bibitem{16}
S.~Li, T.~Azfar, and R.~Ke, ``{Chatsumo: Large language model for automating traffic scenario generation in simulation of urban mobility},'' \emph{IEEE Transactions on Intelligent Vehicles}, 2024.

\bibitem{10}
R.~Khanam and M.~Hussain, ``{Yolov11: An overview of the key architectural enhancements},'' \emph{arXiv preprint:2410.17725}, 2024.

\bibitem{1}
N.~Manakitsa, G.~S. Maraslidis, L.~Moysis, and G.~F. Fragulis, ``{A review of machine learning and deep learning for object detection, semantic segmentation, and human action recognition in machine and robotic vision},'' \emph{Technologies}, vol.~12, no.~2, p.~15, 2024.

\bibitem{3}
H.~Liu, C.~Li, Y.~Li, and Y.~J. Lee, ``{Improved Baselines with Visual Instruction Tuning},'' in \emph{Proceedings of the IEEE/CVF Conference on Computer Vision and Pattern Recognition}, 2024, pp. 26\,296--26\,306.

\end{thebibliography}


\end{document}